\documentclass[12pt,a4paper]{article}

\usepackage[affil-it]{authblk}

\usepackage{fullpage}
\usepackage[parfill]{parskip}

\usepackage[utf8]{inputenc}
 
\usepackage[hidelinks]{hyperref}

\usepackage{graphicx,rotating}

\usepackage[normalem]{ulem}
 
\usepackage{bm}
\usepackage{drs}

\usepackage{gb4e}
\noautomath

\usepackage{latexsym}
\usepackage{amssymb}

\usepackage{textcomp}
\newcommand{\textapprox}{\raisebox{0.5ex}{\texttildelow}}

\usepackage{natbib}
\usepackage{bibentry}
\usepackage{microtype}
\usepackage{adjustbox}

\title{Exploring Probabilistic Soft Logic as a framework for integrating top-down and bottom-up processing of language in a task context}
\author{Johannes Dellert}
\affil{Department of Linguistics, Icall-Research.de Group\\University of Tübingen, Germany}

\date{\footnotesize CoALLa Project Report Ver.~1.0, 2020\\\url{http://purl.org/coalla}}

\begin{document}

\maketitle

\section{Introduction}
This report describes the work which has provided the central architectural 
component of the DFG project CoALLa (Computational Analysis of Learner 
Language). The goal of the project was to explore a novel approach to automating 
the interpretation of learner answers given a task context, as is required from 
a teacher providing feedback to students who completed a reading comprehension 
exercise. The type of output we are aiming to produce has been called 
form-meaning based target hypotheses (FMTHs). An FMTH for a learner answer is a 
normalization which only contains the minimal number of corrections necessary to 
turn the learner answer into a grammatically correct sentence, while at the same 
time being semantically equivalent to the intended answer. The task of automated 
FMTH generation differs from many other tasks in natural language processing in 
requiring some sort of top-down guidance from the semantic level, because the 
task context provides evidence of the learner's intention, and is therefore 
essential for correctly interpreting the learner output. On learner language, 
the lower levels of linguistic analysis (e.g. morphology) provide less reliable 
information than the higher ones.

In contrast, the development of algorithmic solutions for most tasks becomes 
more difficult as we move up through the levels of linguistic description. 
Tokenization is widely considered a solved problem, reliable part-of-speech 
taggers exist for many languages, and a wide range of techniques for 
lemmatization have turned it into not much of a challenge except for the most 
morphologically complex languages. In contrast, dependency parsing is still a 
very active area of research, and semantic parsing has only recently become 
feasible for English with the availability of large-scale meaning banks for 
training. Given this gradient of difficulty, it is only natural that modern NLP 
architectures subscribe to the bottom-up pipeline model, where specialized 
models are trained to perform each analysis step based on the assumption that 
reliable analyses on the lower layers are available. 

This general paradigm has been very successful, but has also been found to show 
weaknesses whenever one of the low-level analyses becomes unreliable, or even 
only feasible by integrating information from higher levels in a top-down 
fashion. A prime example of this is the field of speech recognition, where at 
least a model over plausible word sequences is needed to disambiguate the noisy 
and sparse signal on the acoustic level. But there are also some higher-level 
NLP tasks where the lower layers do not provide enough information for a 
reliable analysis that higher components could build on. For instance, NLP for 
Arabic is faced with the problem that the orthography does not encode short 
vowels, which strongly underspecifies pronunciation and morphological analysis. 
Arabic is notoriously difficult to read aloud for beginners because doing so 
requires both understanding not only the syntactic structure, but in many cases 
also the context of the utterance, e.g. when the reader needs to decide whether 
to enunciate the masculine or feminine form of the possessive suffix, both of 
which are written alike. Systems designed to solve these challenges could be 
seen as providing a partial solution to the type of top-down bottom-up 
integration we need for our purposes, but such architectures are only beginning 
to include language models of morphology and syntax, and the question of how to 
inform an analysis by a given semantic interpretation has remained open.

This report documents the current state of our attempt to close this gap. Beyond 
our use case, the architecture we describe provides a general framework for 
language analysis in any situation where the structures at higher levels of 
linguistic analysis are more reliably predictable than on lower levels.

The structure of this report is as follows. Section 2 introduces
probabilistic soft logic, which we chose as the basis for our
reasoning engine, from the perspective that is most relevant to our
work, i.e. as a logic-based templating language for large graphical
models on which efficient inference is possible.  Section 3 summarizes
our research leading towards our adaptation of Abstract Meaning
Representations (AMRs) for representing both the semantics of learner
answers and the task context. It also includes a survey and assessment
of the current landscape of semantic parsing for German, and the
applicability of recent tools to the language of the classroom as
opposed to newspapers. We find that current AMR parsing technology is
not yet viable for its intended use in our system, and describe our
preliminary solutions which have allowed us to implement our prototype
system. Section 4 then describes our PSL-based CoALLa architecture
both in terms of the underlying design principles and through examples
motivating these design decisions. Section 5 describes the various 
steps we have recently taken to adapt various existing tools and resources for 
use in the architecture, discussing in quite a bit of detail how we are 
addressing the issues of repair candidate generation, morphological analysis, 
and syntactic parsing. The final section summarizes the current state of the 
architecture and provides our views on the most promising avenues for further 
development.

\newpage

\section{Probabilistic Soft Logic}
Probabilistic Soft Logic (PSL) as introduced by \cite{Bach.ea-17}
is a recent formalism from the field of statistical relational
learning (SRL). Unlike other types of machine learning, SRL allows inference
over complex relational structures, which causes results to be interpretable
very transparently in terms of meaningful entities and relations.
This makes it much easier than in other machine learning frameworks
to extract information about the system's internal state, 
e.g. in terms of meaningful contributions to the output.
Like other approaches to statistical relational learning, PSL allows to specify
graphical models over large collections of ground facts, but it is
unique in using a mixture of logical and arithmetic rules, as well as both
inviolable constraints and heuristic weighted rules, all of which can be
templated using grounding, and the solutions to which assign
consistent belief values between 0 and 1 to each atomic ground
fact. In practice, this makes it possible to use PSL as a specification
language wich supports constructs known from logic programming
(a major framework of NLP systems in the era of symbolic approaches),
while at the same time being fully probabilistic, and therefore
compatible with many modern NLP tools. We chose PSL as the basis for our 
architecture because this combination allows us to seamlessly include 
both world knowledge (as encoded in statistical language models) 
and system knowledge (such as grammar rules) into our reasoning.
Unlike previous more expressive approaches to statistical relational learning,
which did not scale well to large inference problems, 
PSL has the advantage of compiling into a type of
random fields on which maximum a-posteriori inference reduces to a
convex optimization problem, enabling efficient optimization
techniques which let it scale to feasible inference over 
hundreds of thousands of variables.

In order to be able to explain how the constraints and rules interact
in our PSL-based architecture, we will introduce some basic elements
of the template language here. Assume that we are dealing with a learner
answer which comes as a sequence of tokens, and would like to reason
about the POS tags which should be assigned to these tokens. The atomic
judgment in this case is the assignment of some part-of-speech category \textit{c} to 
a token \textit{t}. Judgments of this type could be represented by means
of a two-place predicate \textit{Pcat(\textit{t},\textit{c})}, leading to 
grounded atoms such as \textit{Pcat(weiß,ADJ)}, representing that the 
token \textit{weiß} is an adjective (``white''), or \textit{Pcat(weiß,VERB)} 
to represent an analysis of the same token as a verb (``knows''). 
In PSL, such grounded atoms are written with quotes around the constants,
e.g. \texttt{Pcat('weiß','ADJ')}, whereas strings without quotes in argument
positions are interpreted as variables. Adopting the conventions of logic
programming, we always write PSL variables with initial capitalization.
The generic form of our predicate could thus be written \texttt{Pcat(T,C)}
or, for better readability of rules involving the predicate,
\texttt{Pcat(Token,Category)}.

Logical constraints and rules consist of a disjunction of (possibly negated)
non-grounded atoms, but can (often more intuitively) be written using the 
implication symbol \texttt{->} and the conjunction symbol \texttt{\&} 
in the antecendent of the implication.
In PSL syntax, constraints are followed by a dot, whereas rules are preceded 
by the rule weight and a colon. To give a simple example of a logical constraint, 
we will want our toy model of POS tagging to follow the reasoning pattern 
that assigning more belief to a certain category for a token will imply assigning 
less belief to all of the alternative categories. Using the constant inequality 
\texttt{!=} and the negation \texttt{\textapprox}, this can be easily be expressed 
as a logical constraint:

\begin{center}
 \texttt{Pcat(T,C1) \& C1 != C2 -> \textapprox Pcat(T,C2) .}
\end{center}

The main mechanism by which PSL templates a graphical model is to ground the 
rules and constraints of the model against a pre-defined universe of ground 
atoms. This means we can prevent the model from considering all the symbols from 
a given tagset as options for the second arguments \texttt{C1} and \texttt{C2}, 
by only committing actually relevant options to the underlying atom database. In 
our example of the two plausible categories which can be assigned to the token 
\textit{weiß}, we would only commit the two atoms \texttt{Pcat('weiß','ADJ')} 
and \texttt{Pcat('weiß','VERB')} to the database, and would thereby cause the 
two following groundings of our rule to become part of the computation:

\begin{center}
 \texttt{Pcat('weiß','ADJ') \& 'ADJ' != 'VERB' -> \textapprox Pcat('weiß','VERB') .}
 \texttt{Pcat('weiß','VERB') \& 'VERB' != 'ADJ' -> \textapprox Pcat('weiß','ADJ') .}
\end{center}

In many cases, it is more natural to specify constraints directly in terms of 
the belief values of the atoms involved. For instance, the redistribution of 
belief between atoms which represent different tag assignments to the same token 
can much more precisely be specified by stating that the beliefs assigned to those 
atoms should sum to one, much like a probability distribution. This is why the 
PSL templating language also supports arithmetic rules and constraints. The 
syntax for arithmetic rules centers on (in)equations between weighted sums over 
atoms, and crucially, it supports sums of variable numbers of atoms via 
summation variables. Using a summation variable, our constraint on 
\texttt{Pcat(T,C)} atoms can easily be expressed as follows:

\begin{center}
 \texttt{Pcat(T,+C) = 1 .}
\end{center}

Against our example atom database, this will lead to a single grounded rule
instance which exactly enforces our intended constraint:

\begin{center}
 \texttt{Pcat('weiß','ADJ') + Pcat('weiß','VERB') = 1 .}\\
\end{center}

This covers the main features of the PSL templating language as far
as they are needed to understand the description of the CoALLa architecture.
Some more advanced features will be introduced in other sections of this document,
but for the full specification as well as the details of how MAP inference
against such models is implemented, the reader should consult \cite{Bach.ea-17}.

\subsection{PSL infrastructure}

The PSL reference implementation was not initially designed for our type of 
application. Existing applications of PSL involve far fewer predicates and rules 
than are needed to represent multilevel linguistic analyses, whereas the 
number of atoms in the universe has tended to be far larger than in our case. 
The complexity of linguistic models has made it necessary for us to build an 
extensive analysis and debugging infrastructure for large PSL models, the main 
components of which are described in this section.

To understand the dynamics of a PSL model under MAP optimization, it can be 
helpful to conceptualize the ground atoms as moving parts which can move upward 
(higher belief) and downward (lower belief), and the ground rule instances as 
complex spring-like mechanisms which can be put under stress (the degree to 
which they are unsatisfied), and attempt to diminish the stress by exerting 
pressure on the atoms which they connect. Depending on the role of the atom in 
the rule, the arriving signal can be seen as upward and/or downward pressure on 
the atom. For instance, our examples of a logical rule connect the two atoms 
\texttt{Pcat('weiß','ADJ')} and \texttt{Pcat('weiß','VERB')} in such a way that 
they will be put under stress if one of the atoms increases, and react by 
exerting downward pressure on the other atom. The arithmetic rule will do the 
same in both directions, but will additionally exert upward pressure on either 
atom if the belief of the other decreases. From the perspective of a single 
atom, the dynamics can be described in terms of incoming signals from different 
ground rule instances which exert competing upward and downward pressures, and 
MAP inference will tend to set the atom to a belief value which yields to the 
stronger pressures, i.e. the direction demanded by the rules of higher weight.

\begin{sidewaysfigure}
 \includegraphics[width=\textwidth]{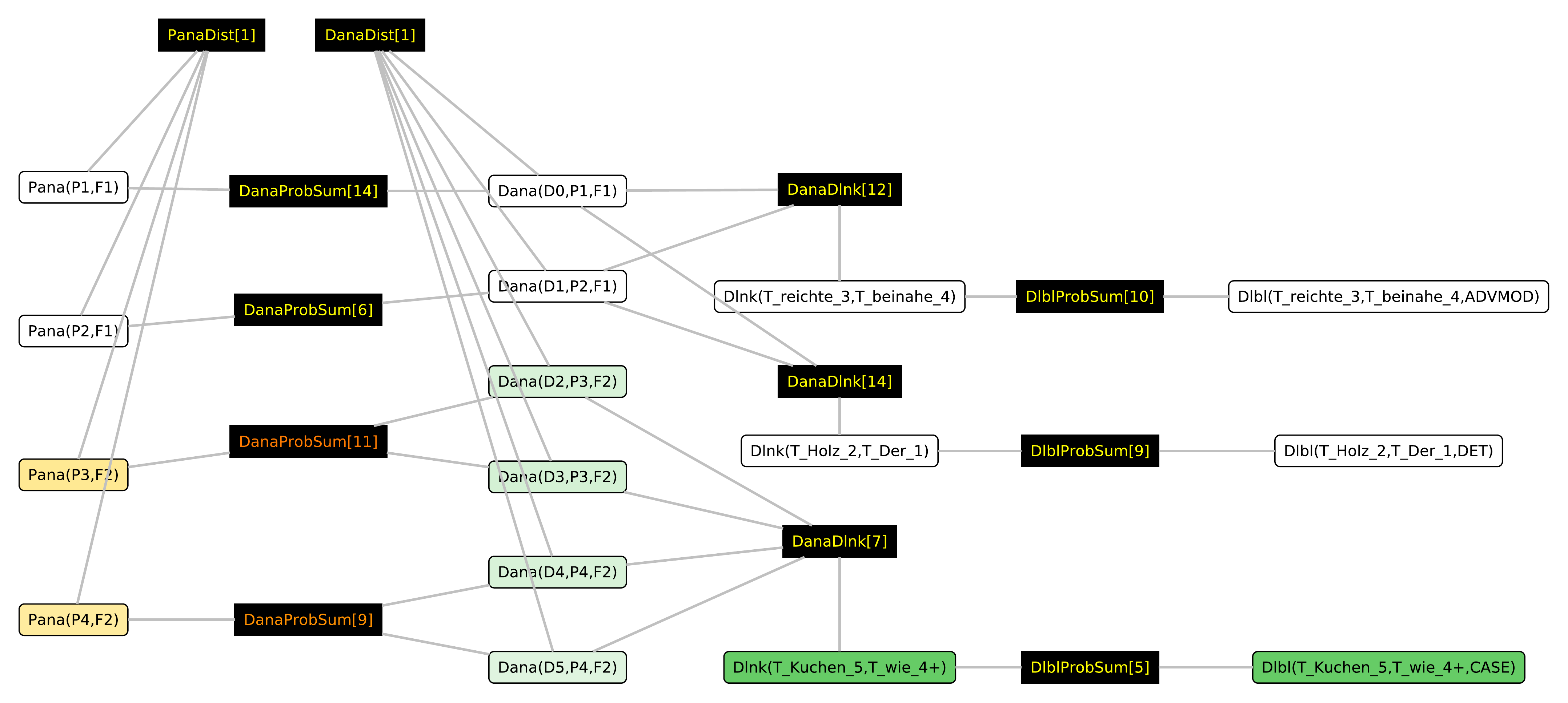}
 \caption{Fragment of a rule-atom graph visualization, showcasing the 
interaction between \texttt{Pana} atoms representing part-of-speech analyses and 
\texttt{Dana} atoms representing dependency structures for the learner sentence 
``\textit{Der Holz reichte beinahe Kuchen.}'' See Section 4 for more details 
about the architecture and the meaning of individual nodes.}
 \label{fig:rag-example}
\end{sidewaysfigure}

In order to facilitate the inspection of this interplay between rules and atoms, 
we took some inspiration from the variable incidence graphs used for 
visualizing the structure of large SAT instances \citep{Sinz-07}. In addition 
to atom nodes corresponding to the variable nodes in variable incidence graphs, 
our rule-atom graphs (RAGs) explicitly model the ground rule instances as a 
second type of node. The graph is bipartite in the sense that all links between 
atom nodes are mediated by rule nodes, and vice versa. As an example of a 
rule-atom graph, a tiny fragment of such a graph arising from our PSL 
architecture is visualized in Figure \ref{fig:rag-example}. In our RAG 
visualization, black nodes represent ground rule instances, and the color of the 
font represent their distances to satisfaction (``stress'') in the MAP result. 
Colored nodes represent ground atoms from different levels of analysis, here 
showing some of the interplay between part-of-speech analyses (prefix P, yellow 
color) and dependency structures (prefix D, green color). The belief assigned to 
each atom in the MAP inference result is visualized by opacity, which implies 
that the white atoms have zero belief assigned to them.

Rule-atom graphs make it much easier to understand the interplay of forces 
acting on the belief values of an atom, but a further challenge of debugging a 
large PSL model is that the interaction chains can range across many 
ground rule instances and atoms, often leading to counterintuitive interactions 
between very different parts of the atom database. For instance, since MAP 
inference amounts to ensuring that as few constraints as possible are 
unsatisfied, belief will often be pushed away from highly constrained parts of 
the model, if the system has the option to avoid committing to a set of atoms 
which would activate the rules in that part of the model. For instance, if a 
learner answer turns out to be difficult to annotate with a dependency analysis, 
an insufficiently constrained PSL model will distribute the belief over many 
part-of-speech sequences in order to avoid having to commit to any single 
dependency analysis. This and similar effects can make it very hard to pinpoint 
why exactly some constraint did not have the intended effects in an inference 
result. One of the obvious possibilities to facilitate debugging in the face of 
such interactions is to only keep parts of the model dynamic while fixing the 
belief values of more distantly connected atoms in place. This will allow the model 
designer to focus first on the interactions within a tightly integrated 
submodel, and then re-connect the parts of the model to test for problematic 
interactions. In order to support such a modular development workflow without 
the need to manually create scaffolding for each debugging task, we invested 
considerable effort into modifying the internal atom database of the PSL 
implementation. This allowed us to add support for temporarily freezing the 
belief values of arbitrary sets of atoms. As expected, this feature made it much 
easier and efficient to fully understand and to refine the behavior of rules and 
constraints.

Towards the end of the project, we started to make use of the fact that many PSL 
rules encode reasoning patterns which are much easier to understand if they are 
stated in natural language. For instance, our example rule could very concisely 
be expressed as ``exactly one part-of-speech must be assigned to each token''.
When exploring our PSL models, we found that when debugging by means of a rule-atom 
graph, it is much easier to follow the reasoning patterns involved if they are 
presented as English sentences. To support this type of output, we created an 
interface which allows to wrap each PSL rule into a Java class which implements 
an interface for generating such verbalizations. PSL rules which are enhanced in 
this way are called \textit{talking rules} within the context of our project. A 
similar interface was added around PSL predicates to create \textit{talking 
predicates}, allowing the inference results to be phrased in natural language as well, 
e.g. ``it is very likely that \textit{weiß} is an adjective in this context''. 
Talking rules and predicates can be combined to automatically generate 
explanations of the type ``\textit{weiß} is unlikely to be a verb here, because 
it is likely to be an adjective, and every token can only have one 
part-of-speech assigned.'' In the final stages of the project, we implemented 
a GUI component which allows us to interactively explore very large rule-atom 
graphs through an interface which is always focused on a single atom. The 
interface verbalizes the meaning of the current atom if it instantiates a 
talking predicate, and lists the verbalizations of all the rules which are 
currently exerting upward or downward pressure on the atom's belief value, 
organized in two blocks which we call the the why block (for upward pressure) 
and the why-not block (for downward pressure). Each rule description contains 
links to the other atoms the corresponding rule node is connecting to, allowing 
a hypertext-style exploration of the system's reasoning when inspecting an 
inference result. For instance, the interface would allow us to click on the 
phrase ``it is likely to be an adjective'' in the previous verbalization, which 
would shift the focus of the explorer to the atom \texttt{Pcat('weiß','ADJ')}, 
explaining the reasoning why this atom received a high belief through 
verbalizations of the ground rule instances acting on that atom.

\subsection{Conventions for PSL predicates describing analyses}

Throughout this document, we will frequently make reference to a variety of PSL 
predicates defined by our architecture. For the sake of readability, we state 
our conventions for predicate names here for reference. All our predicate names 
start with an uppercase letter which is used as a prefix to designate the level 
of analysis the predicate belongs to, followed by a three-letter lowercase 
shorthand representing the exact contents. The nine prefixes currently employed 
by our architecture are the following: \texttt{T} for tokenization, \texttt{F} 
for form variants (target hypotheses), \texttt{P} for part-of-speech tagging, 
\texttt{L} for lemmatization and lexical properties, \texttt{M} for 
morphological features, \texttt{D} for dependency structures, \texttt{S} for 
semantic structures, \texttt{C} for context specification, and \texttt{X} for 
additional helper predicates. To give a few examples, \texttt{Mcas} serves to encode 
morphological case, a \texttt{Dlnk} atom represents a dependency link, and 
\texttt{Sent} atoms express the presence of entities in a semantic model.

During the course of development, we found it necessary to not only talk about 
elementary analysis decisions like single dependency links, but about entire 
analyses on each level of description. The relevant predicates share the suffix 
\texttt{-ana}, making it easy to remember that a \texttt{Dana} atom expresses 
the belief assigned to some dependency analysis, and \texttt{Pana} atoms 
represent different part-of-speech sequences for a given tokenization. The 
interplay between these reified analyses and their elements is a very important
design pattern which will be described much more systematically in Chapter 4.

\subsection{NLP tools as atom and rule generators}

If an analysis on some level of description can be expressed in terms
of atoms that are fed into the PSL model, it makes sense to view external
NLP components as atom generators. For instance, to integrate some external 
POS tagger into our PSL architecture, all we have to do is to write a wrapper 
which extracts the token sequences from our model, calls the tagger on them, 
and then feeds the resulting \texttt{Pana} and \texttt{Pcat} atoms back into 
the atom database. For our example sentence \textit{das Holz roch wie Kuchen} 
(``the wood smelled like cake''), assume that we have a single tokenization 
\texttt{T1}, which is expressed by assigning 100\% belief to the atom \texttt{Tana('T1')}. Our external POS tagger might return a tagging
such as \texttt{das\_DET Holz\_NOUN roch\_VERB wie\_ADP Kuchen\_NOUN}, 
which we would reify as the tag sequence \texttt{P1}, expressed by the 
atom \texttt{Pana('P1','T1')}. The individual tag assignments (which might
be shared across different tag sequences) would include 
\texttt{Pcat('das\_1','DET')}, \texttt{Pcat('Holz\_2','NOUN')}, etc.

Now we have introduced atoms representing POS sequences and individual
tagging decisions to which belief values will be assigned, but without
any rules, the two types of atoms will not be connected. A very direct
strategy for integrating this would be to explicitly write rules connecting
the relevant atoms. For instance, if the tag sequence \texttt{P1} includes
the (erroneous) tagging \texttt{das\_PRON}, we could generate a rule

\begin{center}
 \texttt{Pana('P1','T1') -> Pcat('das\_1','PRON') .}
\end{center}

This would make sure that the model will have to put as least as much
belief on \texttt{Pcat('das\_1','PRON')} as on the tag sequence \texttt{P1}.
The reverse does not hold, because several candidate tag sequences 
will often agree on some (if not most) of the tag assignments. For this
reason, if we assume that \texttt{P1} and \texttt{P3} form
an exhaustive list of all part-of-speech analyses which imply 
\texttt{Pcat('das\_1','PRON')}, the desired interaction can again be
stated much better as an arithmetic rule:

\begin{center}
 \texttt{Pana('P1','T1') + Pana('P3','T1') = Pcat('das\_1','PRON') .}
\end{center}

In a fully integrated module, this will transmit any incoming pressure on the 
atom \texttt{Pcat('das\_1','PRON')}, for instance from a dependency parser which 
has difficulty integrating the word \textit{das} as a pronoun, and distribute it 
equally to the tag sequences which imply this particular tag assignment. The 
interplay of these rules across all assignment decisions will determine which of 
the two part-of-speech sequences gets assigned the higher belief during MAP 
inference. This is a first small example of the way in which top-down guidance 
is implemented in the CoALLa architecture.

\subsection{Binding together larger structures through grounding}

During our exploration of design patterns for large PSL models, we found that 
generating sentence-specific rules as just illustrated is not only error-prone 
due to the need to generate (and then parse) quite complex rule strings, but 
also inefficient due to an implementation which is optimized towards handling 
large number of atoms as opposed to rules. It is therefore generally better to 
avoid turning the output of NLP components into input-specific rules, but to aim 
instead for much leaner interfaces which rely exclusively on generating atoms.

As a way to give more control over arithmetic rule groundings, PSL includes the 
very useful feature of filters that can be attached to summation variables. 
Filters make it possible to condition the way in which summation atoms are 
resolved on the existence of other atoms. We systematically exploit this feature 
in order to write generic rules. Filters are used to configure the groundings of 
these generic rules via helper atoms which are committed to the database in 
addition to the analysis atoms. For our infrastructure, we adopted the 
convention of prefixing the names of helper predicates with \texttt{X}.

To illustrate the usage of \texttt{X} atoms, we will now illustrate how a helper 
predicate can be used to modify the previously described interface to an 
external POS tagger in such a way that it is based entirely on the injection of 
atoms, circumventing the need for our wrapper code to insert any additional 
rules into the PSL model. We express the connections between tag sequences and 
individual tagging decisions through instances of a predicate 
\texttt{Xcat(PA,T,C)}. For instance, \texttt{Xcat('P1','das\_1','PRON')} 
expresses the fact that the part-of-speech analysis \texttt{P1} assigns the 
category PRON to the token \textit{das}. If we modify the wrapper of the POS 
tagger in such a way that it also commits the relevant \texttt{Xcat} atoms to 
the database, independently of the input we only need a single arithmetic rule 
to express the intended interaction between \texttt{Pana} and \texttt{Pcat} 
atoms:

\begin{center}
 \texttt{Pcat(T,C) = Pana(+PA,+TA) . \{PA: Xcat(PA,T,C)\}}
\end{center}

If \texttt{Xcat('P1','das\_1','PRON')} and \texttt{Xcat('P3','das\_1','PRON')} 
are the only two \texttt{Xcat} atoms with these final arguments, this will 
produce exactly the grounding which we had to generate explicitly in the 
previous section. Exploiting PSL's filtering mechanism in this way across all 
layers of analysis has been the key to making the development of the CoALLa
architecture feasible.

\newpage

\section{Enabling top-down guidance through PSL-based\\ semantic matching against target answers}

With the general principles of integrating existing NLP tools into a common PSL 
model in place, the main open issue in the design of the architecture 
was to decide how to model the task context as a collection of PSL atoms. 
The goal is that during MAP inference, matching a semantic interpretation of the 
learner answer against the context must create sufficient pressure on the overall 
constraint system for the signal to be reliably propagated back into the very 
first layers, allowing the system to decide between several target hypotheses.

Second only to the overall architectural design, the most important step on the 
way towards our preliminary solution was to pick a semantics formalism that is 
adequate for the task, both in being representable as a collection of PSL atoms 
connected by helper atoms, and in having some degree of tool support for German. 
This section explains how the issue of representability guided our exploration 
of semantic formalisms, and motivates our decision to adapt Abstract Meaning 
Representations (AMRs) in the light of the current state of semantic parsing 
for German. It then describes how we used existing tools in 
combination with a manual process to annotate a subset of the target answers 
from the CREG corpus with AMRs, and analyzed the requirements of a wide-coverage 
component for robustly translating the constructs which commonly occur in 
learner answers into AMRs. Finally, we discuss how due to the absence of 
available tools which generalize well enough to the learner language domain, we 
made use of our findings in this analysis to guide the design of our prototype 
of a rule-based greedy graph transformation system for translating dependency 
structures into AMRs.

\subsection{Survey of applicable formalisms for semantic parsing}
In an extensive survey of the existing landscape of semantic formalisms and 
tools,  we explored a wide variety of current approaches to semantic 
parsing. A requirement which severely restricted the number of formalisms under 
consideration was that as on the lower analysis levels, semantic representations 
need to be expressible as sets of independently meaningful atomic relations in 
order to enable smooth integration into the PSL model. This excludes the many 
formalisms which rely extensively on variable binding, or are otherwise not 
fully factored out into a single set of relations, such as underspecified 
representations. In the end, our shortlist of candidates consisted of only three 
formalisms for which implementations exist, and which we therefore investigated 
in detail.

Closest to classical formal semantics, the UDepLambda system by 
\cite{Reddy.ea-17} produces $\lambda$-expressions from sentences, via a 
combination of dependency parsing and trainable mappers between tokens and 
lambda expressions as well as from dependency structures to s-expressions 
determining in which order the lambda expressions are composed before being 
reduced by $\beta$-reduction. The main disadvantage of this formalism for our 
infrastructure is that $\lambda$-expressions are difficult to match to each 
other based on partial overlaps, which would force us to perform the matching 
via first-order structures. This would imply a different data format for learner 
utterances (lambda expressions) and contexts (first-order models), and the 
adoption of techniques from the field of first-order model checking for 
performing the matching against context. Model checking is a very 
well-established field, but is tuned towards efficiently deciding whether a 
proposition holds in a given model or not, whereas gradual measures would have 
to be developed for our approach. For a full system, we would moreover need a 
system which generates models from the relevant fragments of reading texts. 
First-order model building is a notoriously difficult problem for which only 
very few tools exist. Even though we previously explored some approaches in this 
direction \citep{Dellert-11}, a lot of challenging groundwork will have to be completed to make this option feasible, which must be left to future work.

Discourse Representation Structures (DRS) as described by \cite{Kamp.Reyle-93} 
are the most expressive and complex formalism currently used for wide-coverage  
semantic parsing. They are very challenging to generate due to incorporating 
scoped negation, existential and universal quantification, as well as variable 
scopes which are difficult to generate correctly using shallow processing 
methods. On the other hand, discourse representation theory is very attractive 
from the theoretical point of view, since it is the only one among the 
formalisms under consideration which systematically covers discourse semantics. 
While being chosen as the formalism for the Groning Meaning Bank and the 
Parallel Meaning Bank (PMB) by \cite{Abzianidze.ea-17}, tool support for 
DRT processing and parsing is rather limited, and the source code for Boxer, a 
highly developed symbolic DRT parser for English, has not been officially 
available for years. On the other hand, a neural semantic parsers for English 
DRS parsing has recently appeared \citep{vanNoord.ea-18a}, causing 
us to include this formalism into our explorations.

The third formalism, Abstract Meaning Representations (AMR), has rapidly been 
gaining popularity during the past years, and provides comparatively wide tool 
support as well as several competing open-source parser implementations. At the 
core, AMRs are directed labeled graphs over events and entities, with special 
edges connecting events and entities to leaf nodes representing the concepts 
they are instantiating. AMRs have the advantage that they consist of locally 
meaningful node and edge configurations that can be modeled independently. 
Another advantage of AMRs is that well-established similarity metrics like 
Smatch \citep{Cai.Knight-13} have developed within the community in the 
context of comparing parser performance against gold-standard data, which makes 
the task of evaluating generated AMRs against the context much simpler than it 
would be for the other formalisms. The main disadvantage of AMRs is their 
limited expressiveness. By default, AMRs do not support scoping for quantifiers 
or negation, reducing them to a bare-bones structure of entities linked by 
relations.

\subsection{Exploration of existing semantic parsers for German}

As the first semantic parser we evaluated, UDepLambda for German was producing 
sensible output, but the lambda expressions it generated turned out to not 
actually use the full expressive power of the formalism. The UDepLambda output 
essentially binds together some constants using predicates, much like AMR, but 
in a less principled fashion with a lot of unnecessary syntactic elements such 
as nested existential quantifiers and conjunctions around what could more easily 
be expressed as a set of relation tuples. The problem is illustrated in Figure \ref{fig:udeplambda} on the basis of a CREG target answer.

\begin{figure}[htbp!]
German: \textit{Im Jahre 1848 gab es in den deutschen Staaten eine demokratische Revolution.}

\medskip

$\lambda x. \exists y. (\exists z. (\exists v.(\exists w.(gab(x_e) \wedge revolution(w_a) \wedge arg_0(w_e,w_a) \wedge demokratische(w_a) \wedge arg_2(x_e,w_a)) \wedge staaten(v_a) \wedge arg_0(v_e,v_a) \wedge deutschen(v_a) \wedge  nmod.in(x_e,v_a)) \wedge es(z_a) \wedge arg_1(x_e,z_a)) \wedge \exists u. (jahre(y_a) \wedge arg_0(y_e,y_a) \wedge 1848(u_a) \wedge nmod(y_e,u_a)) \wedge im(y_a) \wedge dep(x_e,y_a))
$ 

\caption{UDepLambda output (simplified) for a CREG example sentence.}
\label{fig:udeplambda}
\end{figure}

For DRT, we explored the possibility of extending the neural semantic parser by 
\cite{vanNoord.ea-18a} to German for analyzing the CREG data. One obvious issue 
is that the Parallel Meaning Bank, which is the source of our gold training 
data, contains only about 25\% as much gold data for German as for English and 
there was very little silver (partially manually corrected) data for German (as 
of December 2018). However, the main difficulty in applying the neural DRS 
parser to a language other than English is that the PMB DRS concepts are in 
English across the whole corpus. As a result, the neural DRS parser would need 
to both produce the DRS analysis and translate concepts from German to English. 
To emulate the effect we would expect this additional translation step to have 
on the same amount of training data, we shuffled all the concept identifiers in 
the English training and test data, and observed a performance drop from 0.72 to 
0.55 on the English data, confirming that this is indeed a problem.

Because the expected drop in performance caused by translation was further
exacerbated by the sparseness of the training data, results of our neural DRS
parser trained on the German data turned out to be completely unusable, even for
newspaper-style language. Figure \ref{fig:pmb-drs} shows the DRSs for the
sentence ``Anna Politkowskaja wurde ermordet'', i.e. ``Anna Politkovskaya was
murdered''. Not only is the substring ``poli-'' in \textit{policeman} the only
discernible connection between the relation symbols and the contents of the
sentence, but a drinking action and a man named Tom are hallucinated into the
scene, which is plausible given the nature of the training data. Most crucially,
however, the DRS parser does not manage to learn correct nesting structures
between DRSes from the training data, as evidenced by the distribution of
entities and relations to four unconnected boxes, causing some variable
occurrences to be out of scope.

\begin{figure}[htbp!]
German: \textit{Anna Politkowskaja wurde ermordet.}

\medskip

\drs{x, y}{Of(y,$speaker$)\\ Role(x, y)\\ person(x)\\ policeman(y)}
\drs{z}{drink(z)\\ Agent(z,x)\\ Theme(z,v)\\ Time(z,w)}
\drs{w}{TPR(w,$now$)\\ time(w)}
\drs{v}{Name(v,$Tom$)\\ male(v)}

\caption{Example of neural DRS parser output trained on available data for German.}
\label{fig:pmb-drs}
\end{figure}

Our experiences with using neural architectures adapted to German repeated when
we explored the performance of AMREager-DE as described by
\cite{Damonte.Cohen-18}. Figure \ref{fig:amr-examples} juxtaposes the AMRs
produced for our example sentence and its English translation, demonstrating
that while AMREager works for English to some degree, the model trained for
German via machine translation does not yield any reasonable output. We found
this somewhat surprising given that according to the article, performance on
their data as measured by the Smatch score only decreased by about 10 percentage
points when adapting the system to German. This likely indicates that their
trained models for annotation projection do not generalize well to learner
language, and that large amounts of training data tailored to our use case would
be needed to achieve a usable performance level.

\begin{figure}[htbp!]
 \footnotesize
 \begin{minipage}{0.45\textwidth}
   \begin{verbatim}
Input:
"The wood almost smelled like cake."

Output:
(x4 / smell-01
  :ARG0 (x2 / wood)
  :mod (x3 / almost)
  :ARG1 (x6 / cake-01))

Rephrasing:
"The wood almost smells the cake."
  \end{verbatim}
 \end{minipage}
 \begin{minipage}{0.45\textwidth}
  \begin{verbatim}
Input:
"Das Holz roch beinahe wie Kuchen."

Output:
(v3 / roch
    :ARG1-of (v1 / estimate-01)
    :location (v4 / cake))


Rephrasing:
"Presumably, it 'roches' (= smells) inside the cake."
  \end{verbatim}
 \end{minipage}

 \caption{Comparing AMREager outputs for the example sentence.}
 \label{fig:amr-examples}
\end{figure}

While the disappointing results we observed for existing parsers are mainly due
to the absence of large training corpora for German, even for English, where
semantic parsing is a lot more developed, making use of existing systems would
likely not work very well. All state-of-the-art systems are resource-heavy in
the sense that the need a lot of training data in order to show satisfactory
performance, but to our knowledge there currently are no corpora of learner
language annotated with any deep semantic representation that could be learned.
Also, systems are evaluated mostly on newspaper texts or technical documents with
the goal of information extraction and question answering, and are therefore
not designed towards the task of meaning comparison.

In the end, we decided to focus our explorations on AMRs, the least complex of 
the three formalisms, based on the reasoning that performance is likely to 
improve more quickly for parsing a less complex formalism for which several 
competing parsers exist (for English). Also, we expect that unlike in 
information retrieval, where the task is to translate natural-language 
utterances into query expressions in a formal language, the full expressive 
power of DRT will rarely be necessary for representing the semantics involved in 
reading comprehension tasks. Due to the focus on feedback about the form instead of 
evaluation against a formal model of a world, for this task it should be 
sufficient to perform more shallow semantic matching against target answer 
representations.
 
\subsection{Piloting the AMR encoding of target answers}

In order to make it possible to pilot the matching of learner answer and target 
answer AMRs in the context of our architecture, the current state of semantic 
parsing technology made it necessary to deviate from our strategy to create 
interfaces with existing NLP components, and to create a manually annotated 
development dataset instead. For this purpose, we extracted a balanced subset of 
78 questions with 124 associated target answers and a total of 388 learner 
answers from the CREG corpus.

In order to make the process of annotating the target answers with AMRs as 
efficient as possible, we decided to build on existing technology as much 
as we could. In multilingiual semantic parsing, AMRs with English concept names 
are generally accepted as a language-neutral semantic representation, especially 
for practical systems which rely on combining machine translation with AMR 
parsing. Given the state of AMR parsing for German, this encouraged us to first 
translate the target answers into English, and then put the much 
better-performing English AMR parsers to use for generating a first rough 
approximation to the target answer annotations we desired. In an initial 
comparison of system output on a dozen target answers, it quickly became clear 
that on our domain, the CAMR parser presented by \cite{Wang.Xue.ea-2015} 
consistently produces slightly better results than the AMREager system by 
\cite{Damonte.ea-17}. We therefore took the output of CAMR on our English 
translations of the target answers as the basis for our pilot AMR 
annotations.

In a manual post-processing process guided by version 1.2 of the AMR 
specification \citep{Banarescu.ea-14}, we attempted to create the AMRs which a 
hypothetical ideal semantic parser for Standard German should produce. In the 
official specification as well as in the CAMR output, concepts are specified not 
by terms of the original language, but by English lemmas, which are 
disambiguated through cross-references to PropBank frames 
\citep{Kingsbury.Palmer-02} whenever possible. To evaluate and correct the 
mappings produced by CAMR, we relied on the exhaustive list of English aliases 
to frames provided by the PropBank project website\footnote{\url{ 
http://verbs.colorado.edu/propbank/framesets-english-aliases/}}. Throughout the 
detailed discussions of the 124 pilot AMRs within our team of two 
annotators\footnote{Franziska Linnenschmidt and Johannes Dellert}, in every case 
we managed to find a set of PropBank senses and AMR relations which could be 
combined into what we would consider an adequate abstract meaning representation 
for the target answer.

To illustrate the amount of post-processing required, we give an example of
CAMR output and the final AMR annotation for an example target answer in Figure \ref{fig:amr-annotation}. 

\begin{figure}[htbp!]
\centering
 \small
 \begin{minipage}{0.45\textwidth}
   \begin{verbatim}
English translation:

"The light from the oven 
 fell on the child's face."

CAMR Output:
(x6 / fall-01
  :ARG1 (x2 / light
    :source (x5 / oven))
  :location (x11 / face
    :ARG0 (x9 / child)))
  \end{verbatim}
 \end{minipage}
 \begin{minipage}{0.45\textwidth}
  \begin{verbatim}
German original:

"Das Licht vom Ofen 
 fiel auf das Gesicht des Kindes."

Final AMR annotation:
(x6 / illuminate-01 
  :ARG0 (x2 / light 
    :source (x5 / oven)) 
  :ARG1 (x11 / face 
    :poss (x9 / child)))
  \end{verbatim}
 \end{minipage}

 \caption{Example of CAMR output and our final annotation of a target answer.}
 \label{fig:amr-annotation}
\end{figure}

The noun phrase \textit{light from the oven} is translated correctly, but CAMR 
attaches \textit{child} as the first argument to \textit{face}, despite not 
mapping it to a PropBank frame which would license the argument. Presumably due 
to not seeing any instance of light ``falling'' in its training data, CAMR 
defaults to the default frame for \textit{fall}, which according to the 
specification involves downward movement of a thing falling in the role 
\texttt{Arg1-PPT} (patient), and would include the end point of the movement in 
the role \texttt{Arg4-GOL} (destination). CAMR interprets the light as 
undergoing a downward movement not onto, but near the face of the child. A good 
representation of the semantics of this sentence will of course require a 
replacement of the metaphorical use of \textit{fall} by something more concrete. 
In this case, we found the PropBank frame \texttt{illuminate.01} to be adequate 
for representing the semantics, with the causer of the illumation as 
\texttt{Arg0-PAG} (the agent) and the entity illuminated as \texttt{Arg1-GOL} 
(destination). An AMR parser that can produce this mapping would have to learn 
that ``a light X falls on Y'' is semantically equivalent to (and should be 
normalized as) ``X illuminates Y'', which would require many annotated training 
instances, or a handcrafted rule describing this use of the verb \textit{fall}. 
The wrong analysis of the relation between \textit{child} and \textit{face} 
should be easier to fix based on similar training instances, or a generic rule. 
This example will serve to demonstrate the complexity of the AMR mapping task if 
taken seriously, and why we chose to work with a pilot study based on manually 
annotated authentic examples for the time being.

\subsection{Analyzing the requirements for\\ a rule-based dependency-to-AMR mapping}

Our pilot AMR annotations made it possible to investigate the question why 
existing AMR parsers had such difficulty in adequately processing our target 
answers, and more generally to get an estimate of the overall complexity of the 
dependency-to-AMR mapping step. The main thrust of these investigations was 
towards gauging whether it would be feasible to implement a rule-based graph 
transformation approach that could perform well enough on dependency analyses of 
learner answers.

As her bachelor thesis project, \cite{Linnenschmidt-20} recently started to 
investigate this question by extracting path mapping statistics between 
universal dependency analyses and the AMR annotations for our development set of 
target answers. Her investigations were focused on automatically determining and 
keeping a tally of the paths through the AMR which correspond to different link 
types in the dependency parser output. The algorithmic solution proceeds as 
follows. While traversing the dependency tree from the root, the lemma assigned 
to each newly encountered node is mapped to one of the concepts of the AMR based 
on a greedy alignment strategy on the basis of a medium-coverage database of 
many-to-many translational equivalents between English and German content words. 
If either or both of the two lemmas on a dependency link could not be mapped to 
an AMR concept, the dependency link is counted as having been deleted during the 
transformation. Otherwise, a breadth-first search is performed within the AMR to 
find a shortest path connecting the two concepts to which the relevant lemma 
pair was mapped. The resulting path mappings can be seen as a close 
approximation to the overall dependency-to-AMR transformation process which 
would have to be automatized to implement the bridge between the two structures.

The resulting 1,106 path mappings provided us with enough data to derive 
estimates of the complexity of the mapping task for each dependency label, by 
counting the number of distinct AMR path patterns which resulted from mapping 
the dependency links with the relevant label to equivalent paths in our AMR 
annotations. Abstracting over the lemmas and concepts involved, 272 patterns of 
mappings from dependency labels to AMR relation sequences  would be needed to 
describe the entire transformation process for our 124 AMRs, covering a wide 
range of syntactic constructions and semantic phenomena. One of these patterns 
would be the mapping \textit{nummod} $\mapsto$ (\textit{unit}, \textit{quant}), 
which is instantiated by the transformation of the dependency link \textit{Jahr} 
$\rightarrow$ \textit{vier} into a partial AMR that can be representd as 
\texttt{(x1 / temporal-quantity :unit year quant: 4)}. The limited number of 
these mappings encouraged us to pursue a rule-based greedy graph transformation 
approach for our pilot study, as described in the next subsection.

The challenge for such a rule-based implementation of dependency-to-AMR mapping 
will be to determine reasonable conditions for the application of each variant. 
For instance, our example pattern should not actually be applied to each 
\texttt{nummod} link (for which five different patterns exist in our data), but 
only to those links which attach cardinal numbers as dependents to heads with 
the lemma \textit{Jahr}. Given our preliminary data, we were able to estimate 
the complexity of the task of defining such conditions for each case by 
computing the entropy of the distribution of relation sequences that each link 
type is mapped onto. For instance, the 46 \textit{amod} links were mapped to 15 
different relation sequences, but 18 of them simply mapped to \texttt{mod} 
relations, and 8 to \texttt{ARG1-of} relations, while 10 of the 15 sequences 
only occurred once. The resulting entropy value is 2.98, i.e. we would expect to 
minimally need about 3 bits of information to make the perfect decision. 
Comparing this number across dependency labels, despite the currently still low 
quality of our dependency parses we found clear differences in the 
predictability of the relation sequences. For instance, the relation sequency 
distribution for the label \textit{nsubj} (nominal subject) had an entropy of 
3.59, whereas the corresponding value for the label \textit{obl} (oblique 
argument) was 5.31, indicating, perhaps not surprisingly, that the semantic 
relations corresponding to oblique arguments will be much more varied, and more 
difficult to implement, than the relations corresponding to nominal subjects.

\subsection{Prototyping greedy dependency-to-AMR transformation}

For the prototype of our dependency-to-AMR mapping component, we were able to 
stay within the framework of treating both dependency structures and AMRs as 
labeled directed graphs over different symbol sets. From this perspective, the 
translation amounts to relabeling nodes and links, deleting nodes, and rewiring 
links, all of which need to be supported by our rewrite rule format. Since the 
features of our system are likely not final yet, we will refrain from providing 
a full formal definition of our rewriting mechanism here, but instead illustrate 
the capabilities of the current system using examples of the rule format.

As a first example, take our default rule for translating \textit{obj} links into \texttt{ARG1} relations. The notation in our rule file format is as follows:
\begin{center}
\texttt{V, O, V --OBJ-> O => V, O, V --ARG1-> O;}
\end{center}
Essentially, the left-hand side of each rule (before the \texttt{=>}) specifies 
the configuration of nodes and links in the dependency structure to which the 
rewrite rule applies. In this case, all that is needed are two nodes connected 
by a link with the dependency label \textit{obj}. The node representations are 
variables, whose names can be chosen mnemonically within the context of a rule. 
The right-hand side of the rule, describing the result of the transformation on 
the AMR side, states that both nodes remain in the structure, and that the link 
between them is relabeled to the AMR relation \texttt{ARG1}.

To understand why all nodes need to be mentioned on both sides of the rule, 
consider the generic rule for deleting articles (assuming that other determiners 
such as quantifiers are covered by more specific rules):
\begin{center}
\texttt{A, B, A --DET-> B => A;}
\end{center}
This rule matches pairs of nodes connected by \textit{det} links, and deletes both 
the dependent and the link by not reproducing it on the right-hand side of the rule.

As a final example in order to illustrate the more detailed pattern matching 
necessary for many more complex constructions, consider the rule which maps the 
German verb \textit{riechen} together with an oblique argument marked by the 
preposition \textit{nach} or \textit{wie}, to the PropBank frame corresponding 
to the English construction \textit{to smell of something}:
\begin{center}
\texttt{
V[cat=VERB,lem=riechen], 
V --NSUBJ-> S, V --OBL-> O, O --CAS-> C[lem=nach|wie] 
=> V[cnc=smell-02], S, O, V --ARG1-> S, V --ARG2-> O;}
\end{center}
Here, the match is no longer conditioned only on the presence of some link 
with some label, but on additional features of the nodes involved. Also, the 
rewrite rule transforms a chain of two dependency links at once, and transforms 
it into a single concept node with two argument relations.

In our current implementation, the rules whose syntax we have just shown are 
applied repeatedly to the dependency structure in a greedy fashion, i.e. by a 
linear search through the rule file, always applying the first rule whose 
dependency side fully matches. Our plan for rule selection is to base it on a 
specificity ordering that prioritizes rules which match larger numbers of node 
features and link labels. At the moment, such an ordering is specified 
informally by putting the lemma-specific rules more towards the top of the file, 
and keeping the generic default rules at the bottom. As an alternative to greedy 
rewriting, and more in line with our architecture's support for including 
several candidate analyses on each level of description, we will experiment with 
applying the rules non-deterministically, exploring several paths through the 
space of rewrite sequences and accepting all those which result in fully 
converted AMRs.

While it would be straightforward to implement rewrite rules which provide full 
coverage of our current development set, our goal is of course to widen the 
coverage as much as possible. Until very recently, the main obstacle for putting 
this into practice has been the lack of high-quality digital resources which 
fully specify the mapping from the argument positions of many German verbs onto 
semantic roles, as would be provided e.g. by an equivalent of the PropBank for 
German. Fortunately, the lack of such a resource has recently sparked a strand 
of research which attempts to combine analyses of parallel corpora with 
high-quality English-specific resources in order to leverage the work done for 
English for the creation of high-coverage resources for other languages. For our 
specific usage scenario, the most promising project in this direction seem to be 
the Universal Proposition Banks by \cite{Akbik.ea-15}. For instance, the 
\textit{riechen} entry in Universal Propositions 
German\footnote{\url{
http://alanakbik.github.io/UniversalPropositions_German/riechen.html}} includes 
a roleset \texttt{riechen.02} which is mapped to the desired PropBank frame 
\texttt{smell.02}, and contains three examples which provide the information 
that the \texttt{ARG2} role can be realized as a prepositional phrase headed by 
either \texttt{nach} or \texttt{wie}. In our example case, it would thus be 
possible to automatically generate our hand-written AMR conversion rule from the 
information contained in the Universal Proposition Bank for German. It is 
therefore going to be the main resource for our efforts at widening the coverage 
of our rule-based AMR conversion component.

\section{CoALLa architecture}
Building on our PSL debugging infrastructure, 
we were able to rapidly develop and test sets of predicates for expressing the
relevant structures on various levels of description. After arriving at a
consistent solution for representing all relevant structures as
collections of PSL atoms bound together by rules, it became possible able to
start integrating several stand-alone analysis components for Standard
German into a first prototype of the architecture. In addition,
several classes of cross-layer implicational rules were developed to
express a range of hard constraints (such as governing and agreement
phenomena between words linked by certain types of dependency
links). Because they are implemented as disjunctions in a fuzzy
logic, such rules support both deductive and abductive inference
patterns out of the box. For instance, an implicational grammar rule
like ``if an adjective modifies a noun, it should agree with its head
in number, case, and gender'' will operate in a deductive fashion,
which will allow the detection of agreement errors in learner
language, but also trigger abductive reasoning patterns, causing the
system to instead revise the decision to link an adjective to a noun
if there is too much counterevidence, e.g. if there is not only a lack
of agreement, but other rules applying to the adjective (e.g. its
position relative to the head noun) are not satisfied either.

Perhaps not surprisingly, integration of the different components and their 
interaction has turned out to be very challenging, causing the architecture to 
go through several iterations. For instance, our initial attempt to directly 
enforce dependency structure axioms in PSL was found to lead to a large number 
of only weakly constraining rules, which represented a very flat belief
distribution over possible dependency structures at high inference costs. This 
is one of the instances in which we decided to reify entire structures in order 
to constrain the optimization problem more tightly, and several similar complete 
redesigns were necessary to finally arrive at a promising prototype. The current 
iteration of our prototype is the first in which all levels of description are 
fully integrated, although the state of current semantic parsing technology for 
German (see previous section) has made it necessary to build on our a 
proof-of-concept rule-based graph transformation module for the time being.

The architecture of our current prototype integrates analyses of the learner 
input on eight layers, some sequence-oriented and some hierarchical, with 
different analyses binding together elementary local structures, such as the 
assignment of a POS tag (on the P level), the assignment of a head in a 
dependency structure (on the D level), or the assignment of an entity to a 
specific role in a frame (on the S level). On each layer, the belief values are 
constrained to represent a probability distribution over a limited number of 
lower-level structure candidates, whereas linguistically motivated rules can 
refer to the individual parts of each analysis in order to share information 
between analysis layers. For instance, the fact that German proper nouns do not 
carry determiners, which is difficult to enforce in a statistical dependency 
parser, can be expressed by the rather intuitive grammar rule \texttt{Dlnk(H,D) 
\& Pcat(H,'PNOUN') -> !Pcat(D,'DET')} (``a dependent of a head with UD tag 
\texttt{PNOUN} should not have the UD tag \texttt{DET}''). As explained in 
Section 2, PSL allows such rules to either be strict constraints that will not 
be violated, or heuristic rules with learnable weights. We view this distinction 
as a framework for integrating rules in the tradition of symbolic grammar 
engineering with modern statistical language models, while still remaining 
tractable as long as we focus on a limited number of candidate analyses on each 
layer. In our prototype, this is achieved in a straightforward manner by relying 
on beam search. Currently, the different candidate analyses are not weighted
by their quality, but weighted rules could easily be added to configure
prior beliefs about these candidate analyses, in case the need arises on
our way towards wider coverage.

\begin{figure}[htbp!]
\centering
\includegraphics[width=\textwidth]{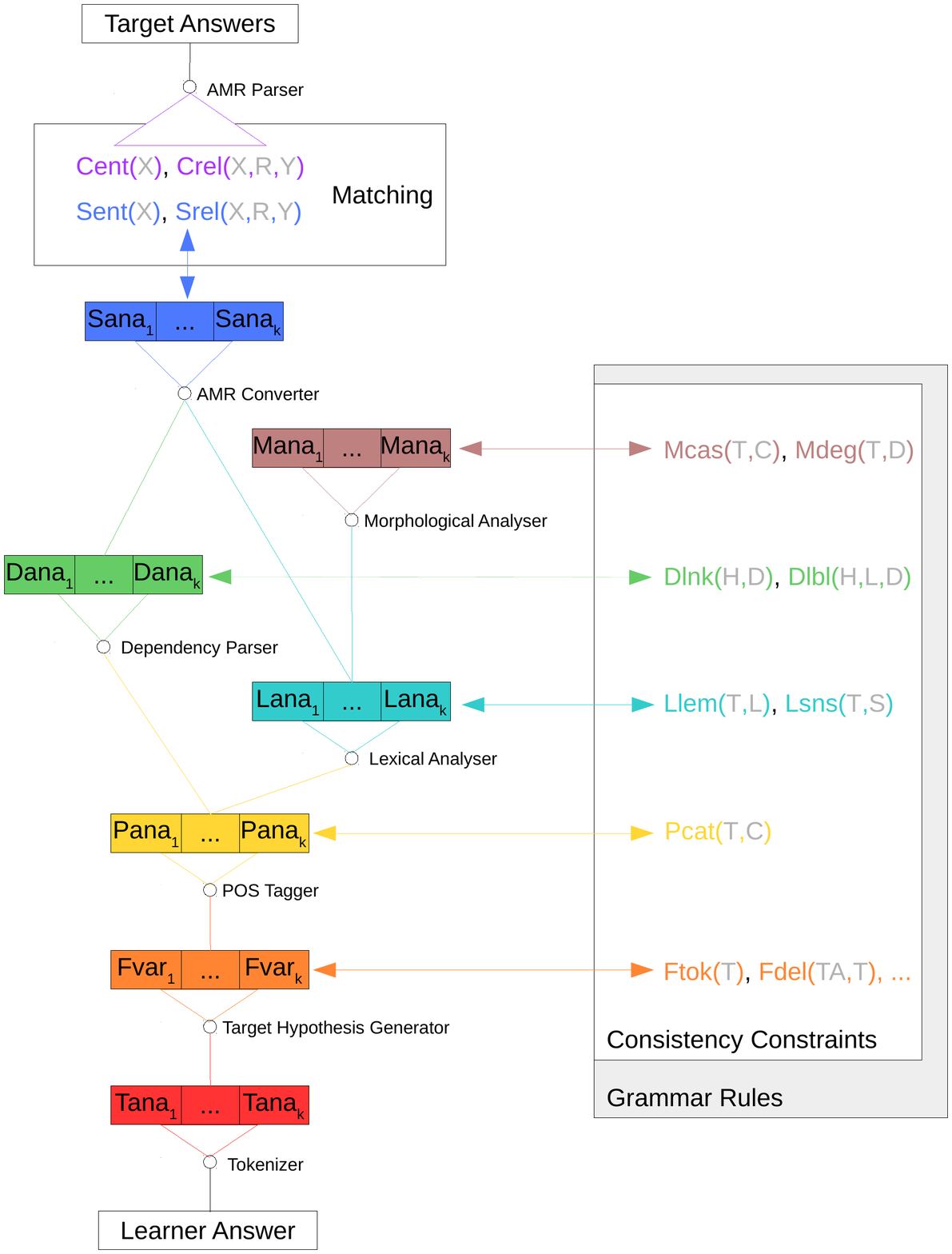}
\caption{Sketch of the overall system architecture.}
\label{fig:architecture}
\end{figure}

Figure \ref{fig:architecture} shows a schematic representation of the prototype architecture. 
As in the discussion of PSL design patterns, reified analyses for each layer are 
represented by PSL predicates with the suffix \texttt{-ana}, e.g. \texttt{Pana} 
for part-of-speech analyses, and \texttt{Dana} for dependency structures. The 
belief values for the analysis atoms on each layer are constrained to form 
probability distributions. Existing NLP tools which support output of the k-best 
analyses are used as idea generators, giving rise to PSL atoms representing each 
output variant, and binding together each low-level analysis with the compatible 
higher-level analyses. The interaction between the belief distributions 
over analyses on different layers is implemented by constraints over the 
marginals, enforcing e.g. that the belief mass assigned to a certain 
part-of-speech sequence must not exceed the sum of the belief values assigned to 
all the dependency structures building on that sequence.

The left-hand side of the architecture graph shows how the different analysis
components are coupled together in this way. On the lowest layer, the
architecture considers different possible tokenizations of the learner input.
The second layer, much more crucial to our architecture, is not equivalent to
any of the standard steps of an NLP pipeline, but is responsible for modeling
the normalization of the learner input. Each of the possible normalizations is
represented by an \texttt{Fvar} atom, which is tied together with atoms
describing the normalization steps, e.g. token insertions and replacements. 
Many possible strategies for target hypothesis generation could be employed here, 
and the systematic exploration of combinations of existing form normalization
methods remains a focus of ongoing work.

Each target hypothesis is run through a k-best part-of-speech tagger, leading to
different part-of-speech sequences, each of which is made up by a set of
individual annotation decisions represented by atoms of the form
\texttt{Pcat(Token,Category)}. This predicate may serve again to illustrate the
design pattern by which information sharing between different analyses is
implemented in our architecture. As described in Section 2.4, the belief assigned to a
\texttt{Pcat(Token,Category)} atom is constrained to equal the sum of the belief
assigned to all \texttt{Pana} atoms representing part-of-speech analyses where
the token is assigned to that category. Using a helper predicate
\texttt{Xcat(P,Token,Category)} which expresses which annotations are part of a
POS analysis \texttt{P}, such rules can naturally be expressed in PSL, as our
example demonstrates:
\begin{center}
 \texttt{Pana(+P) = Pcat(T,C) \{P: Xcat(P,T,C)\}}
\end{center}

This type of constraint is used to bind together reified analyses on other
layers of description with their elementary judgments, often representing
linguistic annotations. For instance, every dependency analysis is connected in
this way with a collection of \texttt{Dlnk} and \texttt{Dlbl} atoms each
representing a dependency arc or label. 

In the architecture graph, the types of atoms each analysis is composed of are
given on the right-hand side. These atoms can be used for expressing
linguistically informed constraints (such as grammar rules) binding together
several layers of description, e.g. to enforce agreement or subcategorization
patterns. The values assigned to such atoms in the optimal solution of the
resulting PSL problems sometimes aggregates information that is distributed
across many analyses, e.g. when the system cannot decide between a range of
dependency structures, all of which do, however, share a certain link, which
would then still receive a high belief value, despite the low values assigned to
the full analyses.

The main source of information which ends up being propagated across the layers 
to arrive at an optimal analysis is the final component, where the candidate 
AMRs generated for the learner answers are matched against the AMRs representing 
the target answers. For this matching, the learner answer AMRs are broken down 
into atomic statements of the forms \texttt{Sent(Concept)}, representing the 
existence of an entity belonging to the \texttt{Concept} in the AMR, and 
\texttt{Srel(X,R,Y)} for representing a relation with label \texttt{R} between 
instances of concepts \texttt{X} and \texttt{Y}. The context AMRs are decomposed 
into \texttt{Cent} and \texttt{Crel} atoms in an analogous fashion, and the 
maximal match between learner answer (correction) and target answer is enforced 
by the very simple PSL constraints \texttt{Cent(X) -> Sent(X)} and 
\texttt{Crel(X,R,Y) -> Srel(X,R,Y)}. Much more complex approaches to matching 
the semantic representations are conceivable, but this simplistic approach was 
so far found to exert just the right amount of pressure on the entire PSL 
problem for MAP inference to converge to sensible analyses of our development 
examples.

The current version of our prototype already demonstrates several of the 
envisioned advantages over less complex architectures. Due to the reification, 
existing NLP components for each layer of analysis can be plugged in as 
hypothesis generators, which will allow the system to profit from future 
advances e.g. in semantic parsing for German, or in the quality of low-level 
hypothesis generation modules, which we will discuss in more detail in the next 
section. Also, the architecture remains scalable by focusing on the top-k 
analyses on each layer, where each $k$ can be modified depending on the 
available computing resources as well as performance requirements. Most 
importantly, the architecture does provide the intended flexibility to add 
additional linguistically motivated constraints and rules to the model at will, 
allowing to integrate meaningful grammar rules into the model, whose violation 
by a learner answer can then be detected. With the finalization of this 
architecture at the end of the CoALLa project, we have demonstrated 
that global optimization of a PSL problem which integrates all levels of analysis 
is strong enough to implement top-down guidance in interpreting learner answers.
A mismatch between the highest two layers, i.e. atomic semantic representations of learner answer variants and the target answer providing the context, is propagated
far enough to decide among a number of target hypotheses on the surface form level,
which implies that we have a feasible architecture for automated FMTH generation.

\section{Integration and refinement of\\ low-level analysis components}

During the year since the official end of the CoALLa project, we have started to 
put major efforts into broadening the coverage of the prototype architecture, 
with the goal of fully covering the set of phenomena present in a development 
set of learner answers and FMTHs. The current version of the prototype already 
integrates a newly developed Java wrapper around Zmorge \citep{Sennrich.Kunz-14} 
which is used to build atom generators for tagging, lemmatization and 
morphological analysis, k-best analysis ranking based on token sequence, 
part-of-speech sequence and lemma sequence models derived from the Tatoeba and 
OpenSubtitles corpora, and an arc-factored dependency model trained on the 
Hamburg Dependency Treebank (HDT) as described by \cite{Foth.ea-14}, in the 
Universal Dependencies version which was automatically converted using TrUDucer 
\citep{Hennig.Koehn-17}.

Initially, we found that many existing NLP components and annotated corpora 
which we planned to build on show issues or pecularities which make them costly 
to integrate, or too unreliable for direct use as atom generators. A deployable 
version of the CoALLa architecture will therefore still require quite some work 
focusing on improvements to the individual system components. This section 
reports on the current state of our work in progress as of March 2020, 
describing many of the problems we have been and are still facing, some 
preliminary solutions, and also some ideas on how the remaining problems could 
be tackled based on available resources in the near future.

\subsection{Form variant generation}

The main challenge of form variant generation is to guide it intelligently in 
such a way that the form-meaning target hypothesis will be among the target 
hypothesis candidates generated, while avoiding the combinatorial explosion that 
would result from too simplistic factoring out of options for local edits. For 
instance, we cannot simply explore all token order permutations of the learner 
answer, because the number of form variants which can be assessed as target 
hypotheses in our architecture is limited. Due to the separation of 
responsibilities implemented by the architecture, we do however have the 
convenient situation that any current or future method for generating target 
hypotheses (even the ones which only operate on the surface level) will be easy 
to integrate into the architecture. All that is needed for each such component 
is a thin layer of wrapper code which retrieves token sequences from the atom 
database, and feeds back the target hypotheses building on this input back to 
the PSL model. This allows our work in this area to be encapsulated as a 
separate subproject with the goal of developing target hypothesis generators 
which manage to generate as many of the annotated FMTHs in our development data 
as possible, while still keeping the number of target hypotheses under 
consideration at a level that is still manageable for our architecture. 

The main obstacle to implementing general-purpose statistical or neural methods 
for form variant generation is the scarcity of available training data for 
German. Since unlike in the mainstream of work on grammatical error correction, 
we do not necessarily need our component to be good at suggesting a single 
optimal error correction, but it will be sufficient to generate a wider range of 
plausible target hypothesis candidates, we are primarily exploring an ensemble 
of simple statistical models, each of which is specialized in a single type of 
surface form edits.

For spelling correction, we built a trie containing a large list of possible 
word forms of Standard German, and implemented approximate lookup by assigning 
weights to non-identical lookup paths through the trie. Modifying a standard 
weighted edit distance-based lookup strategy \citep[sec. 
2.3]{Dickinson.Brew.Meurers-13}, we include information from a German word form 
frequency list generated by Hermit 
Dave\footnote{\url{https://github.com/hermitdave/FrequencyWords/}} on the basis 
of the OpenSubtitles corpus \citep{Lison.Tiedemann-16}\footnote{data from 
\url{http://www.opensubtitles.org}}. This is our approach to implementing the 
idea that more frequent word forms are more likely to have been intended by the 
learner than less frequent ones. The frequency information can also be used to 
some degree for context modeling, which would be implemented by inserting with a 
high pseudo-frequency count all the tokens which occurred in the relevant 
reading context. The same idea can also be used to decrease the number of 
legitimate word forms that are outside the form list stored in the trie. The 
weight function can be further tuned by assigning low costs to replacements that 
represent frequent errors due to interference between L1 and L2 orthographies 
(\textit{sh} vs. \textit{sch}) or phonologies (\textit{sch} vs. \textit{ch}), 
much like standard typo correction methods will encode a confusion matrix based 
on the closeness of keys on the keyboard. To give an example, for the input 
token \textit{Fau}, lookup in our default trie (without context information) 
will return \textit{Frau}, \textit{faul}, and \textit{Pfau} as the 
highest-ranked token replacement options. \textit{Pfau} is phonetically closer, 
but in the absence of a context in which peacocks are relevant, the high 
frequency of \textit{Frau} wins out.

As a general solution for creating possible token insertions and deletions, we 
have so far focused on simple probabilistic approaches along the lines of the 
PKU system described by \cite{Zhang.Wang-14}. Starting from the assumption that 
from the perspective of a native language model, errors in the learner answer 
will look like disfluencies, we look for positions of sharply decreasing 
probability in simple bidirectional Markov models of token sequences, 
part-of-speech sequences, and lemma sequences. At each of the apparently 
disfluent positions, the deletion module tests whether deleting a closed-class 
token (such as a preposition or an article) near the given position increases 
fluency, whereas the insertion module will first attempt to find a closed word 
class whose insertion will increase fluency according to the part-of-speech 
sequence model, and then insert the form of that class which fits best according 
to the overall token sequence model. While both modules have turned out to still 
be a bit difficult to tune towards preventing them from being either too 
conservative or overzealous at generating edits, this is not necessarily a 
problem for the CoALLa architecture, because e.g. the variants where a relevant 
preposition was deleted to improve fluency, will provide a less perfect match to 
the target answer, which will discourage those variants during MAP inference 
across all levels. In the future, we plan to expand this approach to generating 
word-order permutations, again testing for the increase of fluency after 
possible local token-swapping operations. We expect that the current emphasis on 
fluency-increasing editing operations might lead to spurious streamlining with 
adverse effects on the quality of target hypothesis if our goal is to produce 
good FMTHs (which, while grammatically correct, can exhibit a limited degree of 
stylistic markedness). Whether this is actually a major problem will only become 
clear during a future full evaluation of our architecture, but we could attempt 
to address it by tuning some parameters of the underlying sequence models, which 
due to the scarcity of annotated in-domain data are currently still of mixed 
quality.

Coming to more complex and general-purpose approaches which might provide an 
even broader range of form variants, \cite{Boyd-18} explored the potential of 
using state-of-the-art approaches to grammatical error correction for English, 
which are building on current methods of machine translation and large amounts 
of training data, for grammatical error correction in German learner language. 
In order to be able to adapt the existing methods to German, she found it 
necessary to join together training data from several sources, including two 
smaller existing error-annotated corpora as well as corrections from edit 
histories from the German Wikipedia, and found that while combining different 
types of annotations did significantly increase the quality of results, the 
output would still have to undergo significant post-processing and filtering to 
become usable as a target hypothesis generation component in the CoALLa 
architecture. Once some effort has gone into these steps as well as into 
increasing the deployability of the resulting software as a standalone tool, the 
results of this work are likely to become a further important building block for 
form variant generation as we shift towards the more complicated options.

While the work on refining and testing our different approaches form variant 
generation continues, for our fully implemented prototype as well as ongoing 
development of the other components of the architecture, we are currently still 
relying on manually encoded sets of candidate target hypotheses for a small 
number of test sentences.

\subsection{UD-compliant POS tagging}

In order to make it as easy as possible for our system to be adapted to other 
languages, we chose to use Universal Dependencies (UD) as our framework for both 
dependency parsing and the underlying morphological analysis components. This 
implies that part-of-speech tagging will only have to assign to each token one 
of 17 the universal POS tags of UD (which we refer to as \textit{UD tags}). 
In a sense, this task is easier than the usual approaches to language-specific 
POS tagging which use more comprehensive tagsets, but the decision also implies 
that most existing wide-coverage tools need wrappers which translate the POS 
annotations into universal POS tags.

While part-of-speech tagging for German is generally considered a solved problem 
with very high accuracy reported by state-of-the-art approaches, these 
well-performing models are based on machine learning from extensive annotated 
corpora of newspaper language, and are typically only evaluated within the 
domain of the data they are trained on. Also, existing tools do not necessarily 
provide the $k$ best POS sequences as output, which implies that these tools 
would have to be heavily modified in order to fully exploit their potential for 
our infrastructure.

Due to their stochastic or even neural nature, modern POS taggers also tend to 
have the problem that while altogether, they do perform better than older 
approaches, the mistakes they make are not limited to picking a different 
logical possibility, but especially on out-of-domain or non-standard data, they 
will sometimes assign a part-of-speech tag that would be erroneous under any 
context for the token in question.

It will certainly be worthwhile to experiment with adapting state-of-the-art POS 
taggers for use in our architecture at some point in the feature. However, the 
good performance achieved by even very simple and resource-efficient HMM-based 
taggers made it appear worthwhile to start by building a simple $k$-best tagger 
in Java, making it possible to generate the PSL atoms directly while computing 
the analysis, instead of a potentially inefficient interface layer which would 
become necessary for external tools written in other languages. Our current 
implementation combines wide-coverage rule-based morphological analysis in order 
to generate the UD tag options for each token, with preliminary sequence models 
and frequency data which can be derived from existing UD-annotated corpora. We 
also experimented with instead basing the tag sequence probabilities on the 
post-processed output of TreeTagger on the Tatoeba corpus, which is closer in 
genre to learner language than the UD corpora of newspaper text, and therefore 
leads to better results on some of our development examples.

For rule-based morphological analysis, we rely on post-processing the output of 
Zmorge \citep{Sennrich.Kunz-14}, which is currently still the best freely 
available computational morphology for German. Zmorge is distributed as a large 
finite-state transducer which was created by combining open-domain lexical data 
from the German Wiktionary with the SMOR morphology created by 
\cite{Schmid.ea-04}. In order to be able to smoothly integrate Zmorge into our 
architecture, we built on a previously developed lightweight Java driver for 
finite-state transducers in HFST 
format\footnote{\url{https://github.com/tdaneyko/jfst}}, converted the newest 
pre-compiled release of 
Zmorge\footnote{\url{https://pub.cl.uzh.ch/users/sennrich/zmorge/}, we use the 
version dated to 2015-03-15.} from the SFST format to HFST, and wrapped the 
resulting transducer into atom generator components for the part-of-speech, 
lemmatization, and morphological analysis layers. We also used this transducer 
to implement a form generator, which we are planning to use as a way of 
generating surface form variants e.g. with different person and tense forms. A 
stand-alone version of the code for lemmatization and form generation has 
already been made publicly 
available\footnote{\url{https://github.com/tdaneyko/zmorge-utils}}, making it 
possible for other researchers to profit from native Java support for Zmorge as 
well.

While most of the part-of-speech categories from the Zmorge output can 
straightforwardly be translated to UD tags, there are some more complicated 
cases where the design principles of the language-specific SMOR tagset used by 
Zmorge deviate from the typologically motivated distinctions in UD. This 
especially concerns pronouns and participles. For instance, both possessive 
determiners and possessive pronouns (in the stricter sense) are tagged as 
\texttt{POSS} by SMOR, whereas UD assigns them either to \texttt{DET} or to 
\texttt{PRON} based on their syntactic functions. More problematically, even 
participles in attribute position are tagged as \texttt{VERB} by UD, which makes 
sense due to the very common  strategy of employing participle constructions for 
subordinate clauses, but is at odds with the syntactic properties of participles 
in German, where this usage of participles is exceedingly rare. This may be the 
reason why even the official German UD corpora do not actually adhere to the UD 
standard in this regard, presumably because they are automatically converted 
from more language-specific annotation schemes as well.

\subsection{UD-compliant morphological analysis}

Our strategy for the more complex task of atom generation on the morphology 
level is very similar to our currently implemented approach to POS tagging. We 
mechanically translate the full analyses generated by Zmorge to UD features in 
order to create an exhaustive list of options for each predicate, and extract 
$k$-best sequences of morphological analyses for each token using beam search on 
lemma sequence and grammar feature sequence models.

A factor which still provides some obstacles to achieving high quality for our 
sequence models is that according to the results of our investigations, none of 
the UD corpora for German actually fully implements the UD specification. As an 
example of the type of problems in the current HDT corpus, the \texttt{Tense} 
feature is supposed to be used to disambiguate the two German subjunctive moods 
(Konjunktiv I and Konjunktiv II), but this distinction is not repesented at all 
in the corpus. Similar problems occur with the participles, which are missing 
the \texttt{VerbForm=Part} feature mandated by the UD standard, even in the 
cases where they were correctly tagged as verbs. Quite a few of these issues can 
be fixed by rule-based and Zmorge-informed post-processing of the HDT 
annotations, and building models on the basis of the resulting more UD-compliant 
annotations, but a few decisions required by UD remain very difficult to model 
stochastically in the absence of very large amounts of correctly annotated data.

While coverage of the lexicon employed by learners in our development set is 
generally very high, increasing the overall coverage of the system will 
necessarily also mean having to expand Zmorge's lexicon. While Zmorge does 
recognize a range of proper nouns, some very common English names (such as Jack, 
Joe, and Tom) are not recognized at all, and the same is true for some very 
high-frequency particles such as \textit{na}, \textit{ok}, \textit{ne}, and 
\textit{he}. Arguably, these rather colloquial forms should not necessarily be 
accepted as good written German by a feedback system for learners, but the 
architecture will at least have to recognize that these forms belong to closed 
word classes, and are therefore good candidates for fluency-increasing deletion. 
Other out-of-vocabulary problems, especially in linguistically complex materials 
lifted from the reading text into learner answers, will likely be fixable by 
ad-hoc expansion of the lexicon based on linguistic analysis of the reading 
text.

A further issue which has been requiring a bit of work is the fact that Zmorge's 
lemmatization follows some conventions which are at odds with the UD standard. 
For instance, all personal pronouns (including e.g. the ones for the first 
person singular) are analyzed as belonging to the lemma \textit{sie} ``she'', 
and the person information is actually encoded by feature values in Zmorge 
output. In UD, \textit{mich} has to be lemmatized to \textit{ich}, so that we 
found it necessary to circumvent and overwrite Zmorge output for almost all 
pronouns in our wrapper class.

All of these issues are far from extraordinary when joining together different 
NLP tools into larger toolchains or architectures, and will only require a 
limited amount of additional development effort to resolve. The same applies to 
the estimation of good frequency distributions for lemmas. To our knowledge, 
there are no freely available fully lemmatized frequency lists for Standard 
German that are derived from large corpora, but an experimental approach to 
distributing the form counts from the OSC-derived frequency list across analyses 
by cross-comparison within Zmorge-generated paradigms has already led to quite 
promising initial results, and we will be happy to share the resulting frequency 
lists with the academic community in the context of a future publication.

\subsection{Robust morphology-free cross-domain dependency parsing}

The dependency parsing layer was the one were we found it most difficult to 
decide whether to attempt to integrate an existing dependency parser for German 
and adapt it to our purposes, or to build a new parser from scratch which more 
exactly fitted our requirements. Given the complexity of the task, it is of 
course unlikely that a new development that arises from just small subproject of 
a project such as ours will perform as well as current state-of-the-art 
technology on standard language. For this reason, we experimentally integrated 
the neural dependency parser by \cite{Chen.Manning-14}, which is distributed as 
part of Stanford CoreNLP\footnote{\url{https://stanfordnlp.github.io/CoreNLP/}}, 
into the architecture, and observed the behavior of the pre-trained German 
models on our development set. The results were surprisingly erratic for 
constructions which would not typically occur in newspaper texts, confirming 
once more our impression that performance of current dependency parsers depends 
much more on the quality and domain relevance of the training data than the 
complexity of the underlying model. In the absence of adequate amounts of 
training data, we concluded that more explicit models whose behavior can be 
tuned in a more finge-grained manner, might still be the better choice.

During our experiments with such dependency parsers on the learner 
answers, we found that while the structure of dependency trees is generally 
inferred quite reliably, inferring correct dependency labels remains much more 
of an issue for German. Due to the free constituent order, high-performance 
parsers for Standard German (e.g. ParZU, \citealt{Sennrich.ea-13}) very much 
rely on a correct morphological analysis, and make good use of case features for 
labeling, whereas dependency parsers without a good underlying morphology 
component (such as the UDPipe, \citealt{Straka.Strakova-17}) tend to suffer 
severe drops in performance when used for cross-domain parsing. Due 
to the nature of learner language, the reliability of the morphological analysis 
that is needed for high-quality label assignment, is unfortunately not given. 
The severity of this problem was already noted by \cite{Ott.Ziai-10}, who 
compared the performance of two dependency parsers on German learner language. 
Both parsers did manage to infer the microstructure correctly, but the labels in 
the macrostructure (including the syntactic roles) were found to be very 
unreliable.

Informed by these findings, we decided to develop our own simple dependency 
parser with very open interfaces for experiments in the direction of 
morphology-free cross-domain dependency parsing. For structure prediction, our 
current parser implements a simple greedy transition strategy which in one pass 
over the input assigns each token to the best head that does not break one of 
the tree constraints, with preferences based on a scoring function. This 
function is computed from a stochastic configuration model with several fallback 
strategies that are parametrized for the categories, lemmas, and relative 
positions of the two relevant tokens, as well as the categories of intervening 
tokens. Labels are assigned in a second iteration, simply picking for each 
dependency link the label with the highest score according to a second 
stochastic model with the same predictors. When trained on the HDT corpus, this 
simple model performed surprisingly well on in-domain data, but when applied to 
our target answers, as might have been expected it displayed the same problems 
at determining the correct macrostructure as state-of-the-art parsers. Extensive 
analysis of the dependent combinations present in the training corpus revealed 
some hints about why this is so. Dependents to nouns only have very few label 
options, and the decision between them can be made based on local information 
only. At the other end of the spectrum, dependent labels to verbs occur in a 
multitude of combinations (thousands of label combinations are attested in the 
HDT UD corpus) that are very hard to distinguish given only the local context.

To summarize, we find that in the absence of morphology restricting the choices, 
the labeling task becomes too complex to be captured by arc-factored models, 
because it is problematic to assign useful scores to individual labeling choices 
when they are actually heavily interdependent. From our perspective, the most 
promising way forward, which our parser implementation will make relatively easy 
to explore in the near future, is to consider the assignment of label 
combinations to all dependents of a head as a single simultaneous operation, and 
to score such simultaneous labeling steps on the basis of a stochastic model 
over label combinations as they occur in the training corpus. Such a label 
combination model should be conditioned on the lemma for those verbs for which a 
lot of training data is available, and follow back to general statistics about 
dependent label configurations for the other verbs.

This strategy could be conceptualized as extracting quantitative data about 
attested valency frames and selectional restrictions from UD-annotated corpora. 
The problem for this strategy will again be that available annotated corpora are 
based almost exclusively on newspaper texts, with heavily distorting effects on 
the usage patterns for many verbs. For instance, there are thousand of instances 
of the verb \textit{geben} ``to give'' in the HDT, but virtually all of them are 
in the \textit{es gibt} ``there is'' construction. As an example of the 
difficulty in extracting selectional restriction data, we can expect 
difficulties in extracting from news texts the information that tea is a 
prototypical object of drinking. Mentions of the every-day activity of tea 
drinking will be much less commonly found in newspapers than mentions of tea as 
a commodity to be grown, processed, bought and sold.

Our plan for attempting to address these problems is to rely on automated 
morphology-informed dependency parsing of non-newspaper corpora which are both 
broader in content and linguistically simpler than the typical newspaper texts 
or technical documents. As in the other cases, the Tatoeba and OpenSubtitles 
corpora are the primary choices for openly available corpora which could make 
this possible. Even if the label combination data that can extracted from 
automated annotations of these corpora turn out to not be sufficient to make 
morphology-free dependency parsing robust, this strategy can be expected to help 
reducing the number of label combinations to consider as options, so that the 
top-down guidance in our architecture will need to distribute belief to fewer 
candidate analyses, which can then more easily be distinguished and filtered by 
the grammar rules implemented as cross-layer PSL constraints.

\section{Conclusion and Outlook}

The previous sections have already covered most of the details of our current 
multi-pronged roadmap on the way towards a wide-coverage system for learner 
language analysis. Across most levels of analysis, the general strategy we have 
adopted so far amounts to an emphasis on continual refinement, which will 
proceed until each of the components becomes good enough four our purposes. As a 
criterion, this translates to limiting the number of spurious analyses inflating 
the size of the overall PSL problems beyond what the implementation can handle.

The semantic level is the one important exception to this general strategy. On 
all of the other layers of the architecture, state-of-the-art tools can be 
emulated well enough, or are performant enough to at least be considered as 
options for inclusion, allowing future improvements to directly be leveraged by 
our system. On the semantic level, we are currently disconnected from potential 
future technological developments. If our architecture is to profit from future 
advances in semantic parsing technology, there will be no way around building on 
a large collection of AMR-annotated German sentences that are at least 
stylistically similar to learner language.

If we want to avoid taking a major detour through an additional project with the 
goal of creating enough data of this type from scratch, an AMR conversion of the 
Parallel Meaning Bank, along the lines of the ARM-to-DRT conversion described by 
\cite{vanNoord.ea-18b}, is the most promising starting 
point in existence. Many of the sentences in the bank are simple example 
sentences from the Tatoeba corpus. The sentences from the PMB, while according to our 
experiments not sufficient in size for training a neural DRT parser, might serve 
as a starting point for bootstrapping a reasonably-sized training set of German 
AMRs for supervised training of current neural systems for the much simpler task 
of AMR parsing.

As soon as the architecture and the basic versions of our atom generator 
components have stabilized, the plan is to release the code for our architecture 
on GitHub. While the main intention behind this step is to foster further 
research in the paradigm that was started by our prototype, we also believe that 
making the code available will be an important contribution to advancing the 
education of computational linguistics students who are interested in applying a 
wide range of linguistic representations in practice, as are needed for 
architectures that support an exchange of information across linguistic levels. 
We hope that this will allow the completion of further successful BA and MA 
thesis projects that can build on a readily available and well-motivated 
architecture, because establishing such an architecture from scratch is far 
beyond what is feasible for isolated student projects, whereas the integration 
or testing of additional atom generator modules is well within the scope of such 
projects.

When it comes to further development of the core architecture, the possibility 
to add grammar rules in the form of constraints ranging over atoms will make it 
straightforward to add mechanisms for reading off meaningful corrections from 
conflicts inside the constraint system. These mechanisms will build on the 
distance-to-satisfaction measure which is used internally by the PSL reasoning 
engine during optimization, and to which our infrastructure provides access. 
This will become the starting point for further research into the direction of 
expanding the CoALLa architecture from an FMTH-generating pipeline to a software 
which can generate meaningful error feedback involving all levels of linguistic 
description.

\bibliographystyle{acl_natbib}
\bibliography{tech-report-psl-bib}

\end{document}